%% file: iclr2027_conference.tex
\documentclass{article} 
\usepackage{iclr2027_conference,times}
\input{math_commands.tex}

\usepackage[hidelinks]{hyperref}
\usepackage{url}
\usepackage{graphicx}
\usepackage{booktabs} 
\usepackage{multirow} 
\usepackage{pifont}
\newcommand{\yes}{\ding{51}}%
\newcommand{\no}{\ding{55}}%

\author{%
\parbox{0.94\textwidth}{%
\raggedright\normalfont
\textbf{Yuehang Li}\textsuperscript{1},
\textbf{Xiao Wang}\textsuperscript{1,*},
\textbf{Qiang Chen}\textsuperscript{1},
\textbf{Yangyi Fan}\textsuperscript{1},
\textbf{Bo Jiang}\textsuperscript{1},
\textbf{Yaowei Wang}\textsuperscript{2,3},
\textbf{Jin Tang}\textsuperscript{1}\\[5pt]
{
\textsuperscript{1}School of Computer Science and Technology,
Anhui University, Hefei, China\\
\textsuperscript{2}Peng Cheng Laboratory, Shenzhen, China\\
\textsuperscript{3}Harbin Institute of Technology, Shenzhen, China.
\\[4pt]}
{\footnotesize\ttfamily
\{yuehangli, e125111017, e126221332\}@stu.ahu.edu.cn\\
\{xiaowang, jiangbo, tangjin\}@ahu.edu.cn\\
wangyw@pcl.ac.cn\\[3pt]}
}}

\iclrfinalcopy 
\begin{document}

\title{Prototype-Guided Dynamic Domain Modulation and a Large-Scale Multi-SKU Benchmark for Shampoo Bottle Surface Defect Detection}

\maketitle

\begin{abstract}
Industrial shampoo-bottle surface defect detection must operate across many stock keeping units (SKUs), where differences in bottle geometry, material, label layout, reflectance, and imaging conditions can substantially alter the appearance of the same defect. Existing industrial defect datasets rarely provide both instance-level defect annotations and product-variant identities, making it difficult to study SKU-aware detection under realistic manufacturing conditions. We introduce Bottle-SKU, a large-scale multi-SKU benchmark for shampoo bottle surface defect detection. Bottle-SKU contains 13,119 images, 20,419 instance-level bounding boxes, seven defect categories, and 48 bottle SKUs, enabling evaluation of both shared defect recognition and SKU-specific appearance variation. To address the challenge of SKU-specific appearance variation, we propose Prototype-Guided Dynamic Domain Modulation (ProDMM). ProDMM constructs a visual prototype for each SKU from training images and predicts a soft SKU distribution for each input. The resulting prototype-conditioned representation dynamically generates low-rank Query/Value modulation for the visual backbone and residual conditioning for defect-text-to-visual attention, while preserving the pretrained Grounding DINO detection pathway. The model is trained jointly with detection, SKU-domain classification, and modulation-regularization objectives. Experiments on Bottle-SKU and the public GC10-DET-SKU metallic surface defect dataset demonstrate the effectiveness of prototype-guided domain conditioning for industrial surface defect detection. 
The source code and model will be released on \url{https://github.com/Event-AHU/Industrial_Anomaly_Detection}. 
\end{abstract}

\section{Introduction}
Industrial visual inspection promises scalable quality control, yet models that perform well in a controlled production setting can degrade when the product, material, illumination, or imaging configuration changes. Reviews of surface-defect detection identify appearance variation, subtle defect boundaries, and strong dependence on acquisition conditions as recurring obstacles to deployment~\cite{bhatt2021surface}. Industrial domain-adaptation studies likewise report that a detector trained on one inspection distribution requires explicit adaptation when transferred to another~\cite{li2023industrialda}. This problem is particularly visible on shampoo-bottle production lines, where one system serves many Stock Keeping Units (SKUs). Changes in bottle geometry, label layout, surface reflectance, viewpoint, and camera elevation can alter the texture, contrast, and geometric context of the same semantic defect.

This deployment problem is more specific than a shortage of labeled images. Adding samples from already observed products can improve interpolation, but it does not by itself establish transfer to a newly introduced SKU. A random image-level split can also place nearly identical product appearances in both training and test sets, thereby measuring interpolation rather than cross-product generalization. The desired detector must therefore separate two factors. Defect semantics should be shared across products, whereas the appearance prior should change with the current product variant. This motivates an evaluation protocol in which product variants are explicit domains and held-out variants are used to test generalization.

Domain adaptation provides an important starting point. Adversarial alignment methods learn representations that make domains difficult to distinguish~\cite{ganin2016dann}. Detection-specific methods align image-level and instance-level features~\cite{chen2018dafa}. Multi-source adaptation combines information from several labeled domains~\cite{yao2021multisource}. These approaches are not a direct solution to the multi-SKU setting. They commonly formulate a source--target transfer problem or seek a domain-invariant representation, while a production line contains many related variants and the appearance differences themselves can provide useful evidence for resolving ambiguous defects. Recent open-set defect recognition also broadens the range of defect appearances that a model can recognize~\cite{Kim_2026_CVPR}, but does not establish an instance-level, SKU-disjoint detection protocol with explicit product-variant conditioning.

Vision-language detectors offer a complementary source of transferable defect semantics. Grounding DINO~\cite{liu2024groundingdino} and subsequent systems use text-conditioned visual interaction to localize concepts beyond a fixed detector vocabulary. Recent work improves this paradigm through large-language-model supervision~\cite{fu2025llmdet,zhou2026led}, retrieval-based open-vocabulary detection~\cite{fu2026wedetect}, semantic augmentation~\cite{cao2026hsadino}, visual-token prompt refinement~\cite{qin2026vitprompt}, richer category descriptions~\cite{yang2026sradet}, and hierarchical objectness calibration~\cite{lee2026hcc}. These methods strengthen the semantic interface between a defect prompt and visual evidence, but the text prompt for ``scratch'' or ``fold'' is still shared across SKUs. Thus, language conditioning alone does not tell the detector how the same defect should appear on a particular product variant, nor does it provide a mechanism for soft transfer among visually related SKUs.

Existing defect benchmarks leave this product-variant setting under-specified. NEU-DET and GC10-DET focus on defects in steel-strip imagery~\cite{song2013neu,lv2020gc10det}, HRIPCB targets PCB defects~\cite{huang2020hripcb}, and MVTec AD emphasizes unsupervised anomaly localization across object and texture categories~\cite{bergmann2019mvtec}. These benchmarks are valuable for measuring defect localization, but they generally do not jointly expose full-product appearance variation, explicit product-variant identities, instance-level defect boxes, and a product-disjoint test protocol. As a result, they cannot directly answer whether a detector can preserve defect semantics while adapting to the appearance of an unseen product variant.
\begin{figure}
    \centering
    \includegraphics[width=1\linewidth]{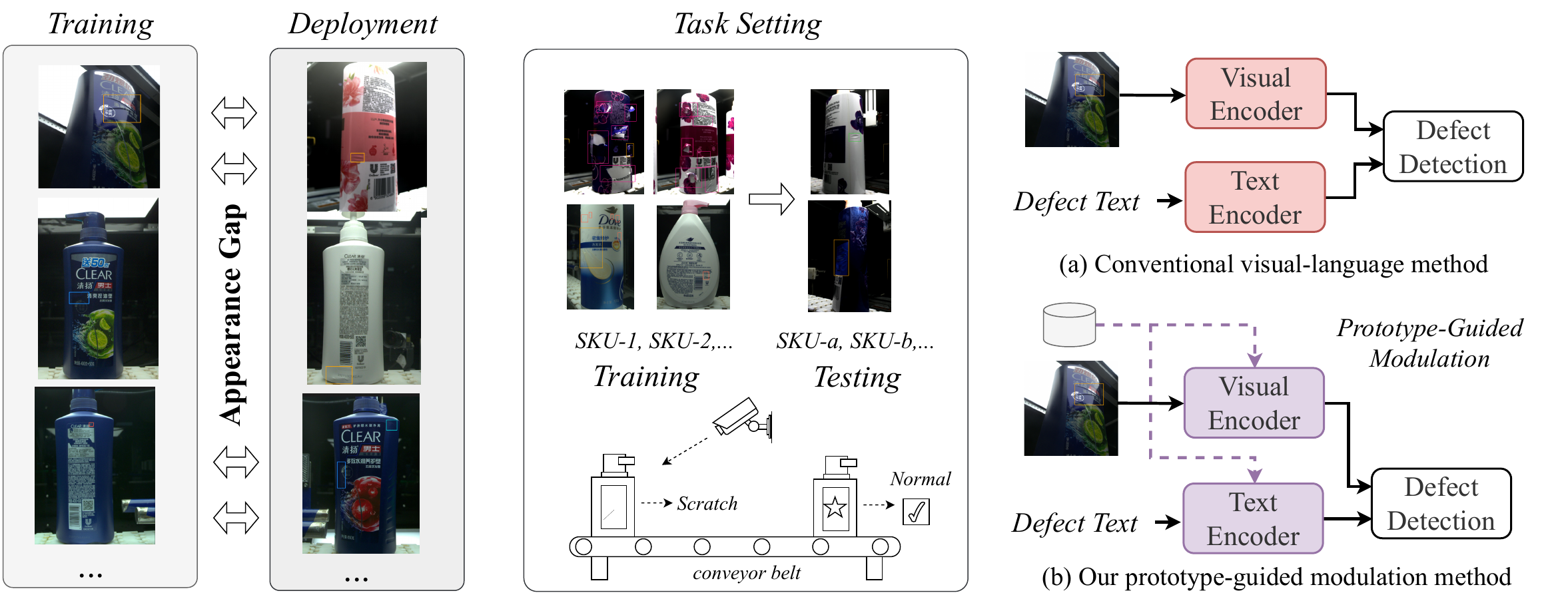}
    \caption{\textbf{Overview of the proposed cross-SKU defect detection 
benchmark and method.} 
(\textbf{Left}) \emph{Appearance gap} between training and deployment. 
Shampoo bottles from different SKUs vary substantially in appearance 
and material, while defects such as scratches and wrinkles remain 
visually similar. 
(\textbf{Middle}) \emph{Task setting}. We adopt an SKU-level deployment 
protocol, training on a set of seen SKUs (\emph{SKU-1, SKU-2, ...}) 
and testing on unseen SKUs (\emph{SKU-a, SKU-b, ...}), which emulates 
the real-world scenario where a production line switches to a new 
product. 
(\textbf{Right}) (a) Conventional visual-language methods feed visual 
and text features directly into the detection head. (b) Our 
prototype-guided modulation injects domain shift into the text and 
visual branches, enabling robust detection on unseen SKUs.}
    \label{fig:first_image}
\end{figure}

To study this problem, we introduce Bottle-SKU, a multi-SKU shampoo-bottle surface-defect benchmark with instance-level annotations and explicit SKU identities, and propose Prototype-Guided Dynamic Domain Modulation (ProDMM). ProDMM extracts a prototype for each training SKU, predicts a soft SKU posterior for an input image, and uses the resulting mixture of prototypes to condition low-rank visual modulation and defect-text cross-attention. This design retains a shared detection pathway for defect semantics while allowing the appearance-dependent computation to vary smoothly across related products. We evaluate the proposed setting on Bottle-SKU and on a pseudo-SKU split of GC10-DET-SKU. Our contributions are threefold:
\begin{itemize}
    \item We introduce Bottle-SKU, a large-scale full-object surface-defect benchmark with explicit SKU identities, professional instance-level annotations, and an SKU-disjoint evaluation protocol.
    \item We propose ProDMM, which converts a soft mixture of SKU prototypes into image-conditioned low-rank QLer/VLer residuals for the frozen visual backbone, preserving the shared pretrained detection pathway.
    \item We introduce Domain Cross Attention (DCA), which injects prototype-conditioned visual evidence into defect-text interaction, and demonstrate the resulting framework on Bottle-SKU and GC10-DET-SKU under product-aware evaluation.
\end{itemize}

\section{Related Work}
\paragraph{Industrial Surface Defect Detection.}
Industrial inspection includes supervised instance detection and unsupervised anomaly localization. NEU-DET~\cite{song2013neu} established a widely used hot-rolled steel benchmark with six visually distinct defect types. GC10-DET~\cite{lv2020gc10det} broadened this setting to ten metallic-surface categories and high-resolution strip images. HRIPCB~\cite{huang2020hripcb} emphasizes fine-grained PCB defects, while MPCG~\cite{tao2018mpcg} studies cover-glass defects under reflective imaging. PVMulti-Defect~\cite{li2023gbhyolov5} and APDDD~\cite{alibaba2023apddd} represent photovoltaic and aluminum-profile inspection, respectively, and MVTec AD~\cite{bergmann2019mvtec} provides a multi-object benchmark for unsupervised anomaly detection. Severstal~\cite{severstal2019steel} contributes large-scale production steel imagery, whereas DAGM~\cite{wieler2007dagm} isolates textured optical inspection with controlled backgrounds. 

In parallel, anomaly methods learn normal appearance rather than enumerating every defect type. PaDiM~\cite{defard2021padim} fits a multivariate Gaussian to deep patch features, PatchCore~\cite{roth2022patchcore} stores representative normal patches in a memory bank for nearest-neighbor scoring, CutPaste~\cite{li2021cutpaste} creates synthetic anomalies through self-supervised transformations, and DRAEM~\cite{zavrtanik2021draem} trains a reconstruction-and-discrimination network on augmented defects. MVTec LOCO AD~\cite{bergmann2020mvtecloco} further tests logical and structural anomalies that cannot be detected from local texture alone. These datasets and methods usually assume a fixed product identity or remove object-level context through local crops. In contrast, our proposed Bottle-SKU dataset evaluates full-object, instance-level detection across explicit SKU identities, exposing a cross-domain generalization problem not addressed by standard benchmark protocols.

\paragraph{Prototype-based adaptation and conditional modulation.}
Prototype-based learning~\cite{snell2017prototypical} and conditional modulation~\cite{perez2018film,yang2019condconv} adapt representations using representative embeddings or auxiliary signals. Domain-adaptation methods~\cite{ganin2016dann,chen2018dafa} align source and target features, while parameter-efficient transfer~\cite{houlsby2019adapter,hu2022lora} injects lightweight trainable modules into frozen backbones. Recent works explore dynamic modulation for domain adaptation and generalization. Feature-level modulation adapts visual features to domain shifts via semantic entropy guidance, channel-wise compensation, content-style decomposition, and concept-aware modulation~\cite{li2025seen,CD-Buffer_2026_CVPR,Guo_2026_CVPRCo-Salient,Sultana_2026_CVPRCoFiDA-M}. Conditional modulation generates input-dependent parameters or adapters to modulate frozen backbones~\cite{Khan_2026_CVPR,Aiello_2025_CVPRDreamCache,zhang2025easycontrol}. Prototype-guided and representation-level modulation adjusts statistical properties or activation patterns for cross-domain robustness~\cite{cai2026Feature-aware,gan2026unleashing,kundu2026boosting}. Unlike these approaches, our Prototype-Guided Dynamic Domain Modulation generates domain shift information. See Appendix~\ref{sec:more-related-work} for a comprehensive review.

\section{Our Proposed Approach} 
\label{sec:ourmethod}
\subsection{Overview}
\begin{figure}
    \centering
    \includegraphics[width=1\linewidth]{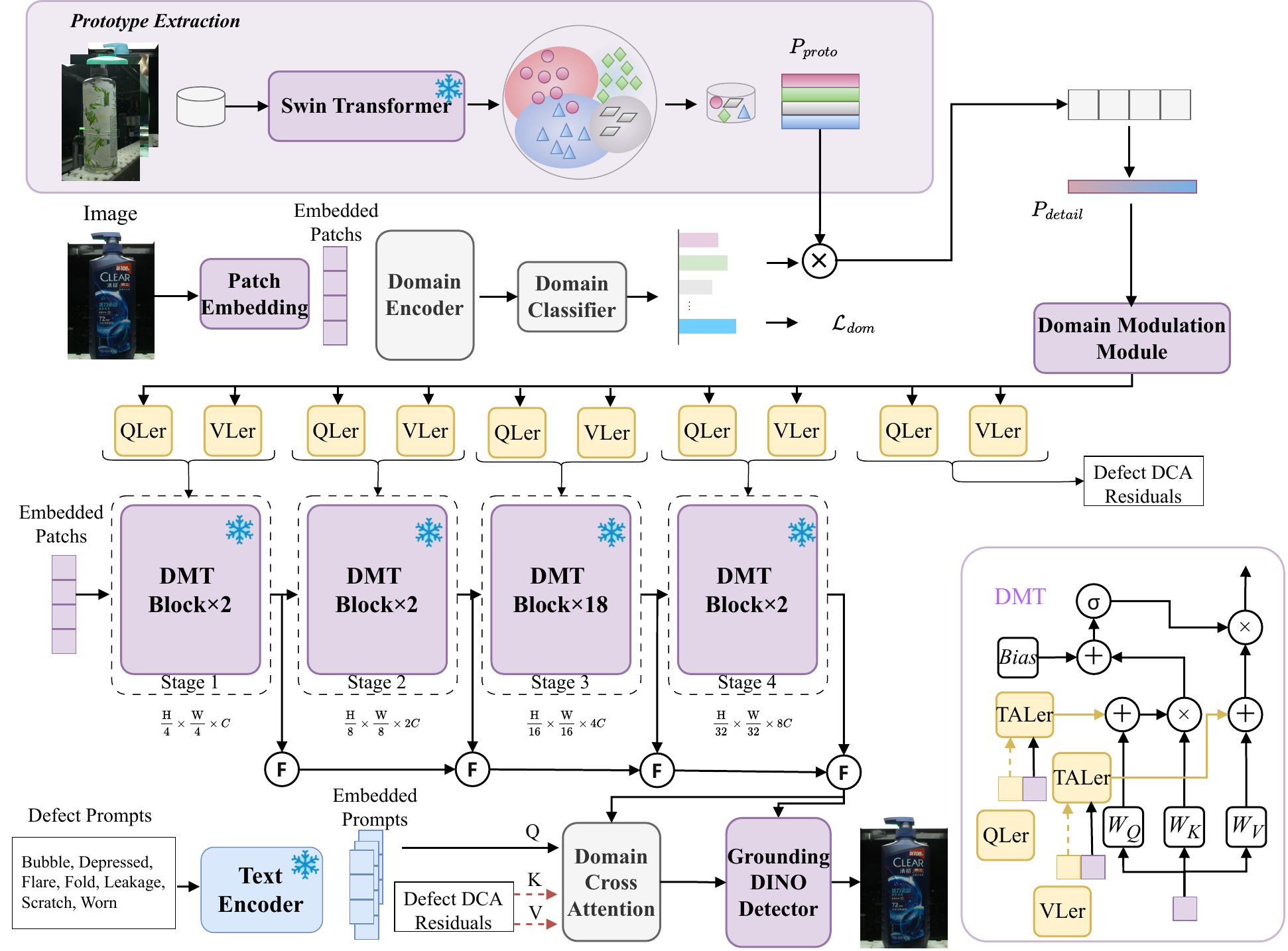}
    \caption{The overall architecture of ProDMM. Offline SKU prototypes condition dynamic QLer/VLer residuals in the Swin backbone and residuals for defect Domain Cross Attention (DCA). DCA updates only the text memory before the original GroundingDINO Feature Enhancer. The bottom-right inset illustrates a DMT block.}
    \label{fig:framework}
\end{figure}

We propose a prototype-conditioned domain modulation framework for multi-SKU defect detection. In the main frozen-backbone setting, the pretrained Swin-B backbone and BERT text encoder are frozen, while the dynamic modulation parameters in DMT, the Domain Cross Attention (DCA), and the GroundingDINO interaction layers are trained. We first construct a fixed SKU prototype matrix by offline. Then, the framework extracts patch tokens of input image and uses a Domain Encoder to predict a soft SKU probability vector. The predicted probabilities softly fuse all SKU prototypes into a domain detail representation, then, this representation drives Domain Modulation Module to generate QLer/VLer modulation coefficients for all Swin-B blocks. The DCA is used to enhance text and visual tokens, then processed by the frozen GroundingDINO Detector to produce defect detection outputs. The whole model is optimized with the detection loss, an SKU-domain classification loss, and regularization terms for the generated modulation parameters. The overall framework is shown in Fig. \ref{fig:framework}.

\subsection{SKU Prototype Construction}
Given input images $I \in \mathbb{R}^{B \times C \times H \times W}$ and defect text prompts, our goal is to perform defect detection and SKU domain classification in a multi-SKU setting. For the $j$-th training image $I_{u,j}$ of the $u$-th SKU, we use the official GroundingDINO Swin-B weights to extract the final visual stage feature for each image to obtain $f_{u,j}$. The SKU prototype is computed as the mean of the image features:
\begin{equation}
P_u = (\frac{1}{M_u}\sum_{j=1}^{M_u}f_{u,j}),
\end{equation}
where $M_u \leq M_{\max}$ at most $M_{\max}$ training images are used for each SKU. Stacking the prototypes of all $U$ SKUs gives the prototype features
\begin{equation}
P_{proto}=
\left[P_1;\ldots;P_u\right],
\end{equation}
where $P_{proto} \in \mathbb{R}^{U \times d_p}$ is computed offline from training images and kept fixed during detector optimization. No image from a held-out test SKU is used to construct this table. The table provides a compact appearance prior while leaving the defect categories shared across SKUs.

\subsection{Prototype-conditioned Domain Modulation}
The Domain Encoder operates on the patch tokens produced by the frozen Swin-B Patch Embedding. The input images is embedded as: 
\begin{align} 
Z_0 = \operatorname{PatchEmbed}(I) \in \mathbb{R}^{B \times N \times d_e},
\end{align}
where $B$ is the batch size, $N$ is the number of patch tokens, and $d_e$ is the patch embedding dimension. A $1\times1$ convolution is then applied to $Z_0$ to extract gate features $G \in \mathbb{R}^{B \times N \times d_g}$. These features are concatenated with $Z_0$ along the channel dimension, yielding $[Z_0; G] \in \mathbb{R}^{B \times N \times (d_e+d_g)}$. A single-layer MLP followed by global average pooling over tokens projects the concatenated features to a domain vector $g$, the Domain Encoder is defined as follows:
\begin{align}
g&=AvgPool(\operatorname{MLP}\left([Z_0;G]\right)), g \in \mathbb{R}^{B\times d_0},\\
h_{dom}&=\operatorname{GELU}(\operatorname{MLP}(g)), h_{dom} \in \mathbb{R}^{B \times d_{dom}},
\end{align}
where $[\cdot;\cdot]$ denotes channel concatenation and $h_{dom}$ is the domain feature. The SKU domain classifier consists of a linear layer followed by softmax and outputs the soft SKU probability $p$:
\begin{align}
p=\operatorname{Classifier}( h_{dom} ), p \in \mathbb{R}^{B \times U}.
\end{align}
Instead of using a hard-selected prototype, ProDMM softly fuses all SKU prototypes using the predicted probabilities, $P_{detail} = p \cdot P_{proto}$, so the model receives a continuous domain representation containing information from all SKU prototypes. Domain Modulation Module (DMM) then maps $P_{detail}$ to the global domain representation $\mathbf{c}$:

\begin{equation}
\mathbf{c}=\operatorname{DMM}(P_{detail}), \mathbf{c} \in \mathbb{R}^{B \times d_{dom}},
\end{equation}
where $\operatorname{DMM}$ is a two-layer MLP with GELU. Thus, all subsequent dynamic parameters are conditioned on global representation $\mathbf{c}$. This soft aggregation retains uncertainty among visually similar SKUs instead of committing to a hard prototype assignment. It is beneficial for inference when it can recouple appearance information from the same product but different appearance.

\subsection{Prototype-Guided Dynamic Modulation with QLer/VLer}
Let $\mathbf{h}$ denote the input tokens of each Swin Transformer block and the block index is omitted below for readability. Since $\mathbf{c}$ is derived from DMM for each image, the corresponding model parameters are computed as follows:
\begin{align}
\theta_Q=\operatorname{QLer}(\mathbf{c}),\qquad
\theta_V=\operatorname{VLer}(\mathbf{c}),
\end{align}
where QLer and VLer are single layer MLPs in implementation. For clarity, we describe $\theta_Q$ and $\theta_V$ as $\theta$. Then, $\theta$ are split to initialize Task Aligned Layer (TALer) \cite{Khan_2026_CVPR} as follows: 
\begin{align}
\{W_{\downarrow}, b_{\downarrow}, W_{\uparrow}, b_{\uparrow}, s\} = \operatorname{Split}(\theta),\\
\Delta=
Scale \cdot \sigma(W_{\downarrow}h+b_{\downarrow}) \odot ( W_{\uparrow}\sigma(W_{\downarrow}h+b_{\downarrow})+b_{\uparrow} ),
\end{align}
where \(W_{\downarrow}\) and \(W_{\uparrow}\) are the down- and up-projection matrices, \(b_{\downarrow}\) and \(b_{\uparrow}\) are the corresponding biases, \(Scale\) is a scaling matrix. The $\theta$ are zero-initialized so that training starts close to the frozen pretrained model.

We obtain $\Delta_{Q}$ and $\Delta_{V}$ from above equation. Let $Q$, $K$, and $V$ be the original attention projections. DMT modifies only Query and Value:

\begin{equation}
\\
\widehat{Q}=Q+\Delta_{Q},\qquad
\widehat{K}=K,\qquad
\widehat{V}=V+\Delta_{V},
\end{equation}
where $\Delta_{Q}$ and $\Delta_{V}$ are produced by the corresponding QLer and VLer. All adapters are conditioned on the same global domain feature $\mathbf{c}$, while different backbone depths learn different appearance-dependent visual relations without introducing a separate detector for every SKU. Keeping $K$ unchanged preserves the pretrained key space and restricts the adaptation to a controlled residual update.

\subsection{Domain Cross Attention}
Let $T$ denote the defect-prompt tokens produced by the text encoder and let $F_{neck}$ denote the concatenated original neck tokens. In DCA, $T$ acts as the query, while $F_{neck}$ provides the keys and values. Unlike freezing the backbone network as a hypernetwork to generate model parameters for modulating modules, the DMM applies QLer and VLer to the same global condition to generate domain shift residuals for the DCA Query and Value:
\begin{equation}
\delta_{Q}=\text{QLer}(\mathbf{c}),\quad
\delta_{V}=\text{VLer}(\mathbf{c}),
\end{equation}
Therefore, the query, key and value are computed as follows: 
\begin{equation}
\begin{aligned}
\hat{Q} = T W_Q + \delta_{Q},\quad
\hat{K} = F_{neck} W_K, \quad
\hat{V} = F_{neck} W_V + \delta_{V}.
\end{aligned}
\end{equation}
The residual vectors are broadcast over the corresponding text or visual token dimension. Thus, QLer/VLer change the product-conditioned content of the Query and Value projections without changing the source sequence used to form the keys. The DCA output is computed by multi-head attention:
\begin{equation}
\operatorname{DCA}(T,V)=
\operatorname{softmax}\left(
\frac{\hat{Q} \hat{K}^\top}{\sqrt{d}} +\hat{b}
\right)
\hat{V},
\end{equation}
where $b$ is a learnable SKU-specific bias. DCA allows defect-text tokens to retrieve image evidence from the multi-level visual representation. This design introduces SKU-conditioned visual evidence into defect-text interaction without replacing the pretrained visual processing pathway. Consequently, the model can preserve shared defect semantics while adapting the text-visual interaction to product-dependent appearance changes.

\subsection{Detection Head and Training Objectives}
The resulting visual features and DCA-conditioned text memory are processed
by the GroundingDINO detection pathway to produce class scores and bounding
boxes. The native detection objective is denoted by $\mathcal{L}_{det}$. For a ground-truth SKU label $y$, domain classifier is supervised by
\begin{equation}
\mathcal{L}_{dom}=\operatorname{CE}(p,y).
\end{equation}
We use one regularizer for the generated DMT parameters, DCA residuals and DCA learnable parameters:
\begin{equation}
\mathcal{L}_{reg}=
\operatorname{Mean}(\widehat\theta^2)+
\operatorname{Mean}(\hat\delta^2)+
\operatorname{Mean}(\hat{b}^{2}),
\end{equation}
where $\widehat\theta$ denotes the generated DMT parameter and
$\hat\delta$ denotes the DCA Q/V residuals.
This regularizer constrains the magnitude of the conditional updates and
helps preserve the pretrained representation. The final objective is
\begin{equation}
\mathcal{L}=\mathcal{L}_{det}+
\beta\mathcal{L}_{dom}+
\gamma\mathcal{L}_{reg}.
\end{equation}
The SKU loss makes the product-appearance code discriminative enough to
produce meaningful prototype mixtures, while the modulation regularizer
limits the magnitude of the conditional updates. Together with
$\mathcal{L}_{det}$, these objectives preserve shared defect
semantics while allowing controlled adaptation to SKU-dependent appearance.

\section{The Bottle-SKU Dataset}
\label{sec:bottle-sku-dataset}
Bottle-SKU contains images of shampoo bottles captured on factory production lines under real-world inspection conditions. The dataset will be publicly released upon publication to facilitate research on SKU-aware surface defect detection.
Table~\ref{datasetlist} compares our Bottle-SKU dataset with existing public surface defect detection datasets. While prior datasets such as NEU-DET, GC10-DET, and MT focus on simple industrial surfaces with low appearance complexity and mostly local inspection, our dataset targets shampoo-bottle surfaces with high appearance complexity and full-object inspection. Compared with conventional defect datasets that focus on a single product or a small number of fixed appearances, bottles in the dataset vary in brand, label design, net weight, viewing side, viewing angle, and camera elevation. These variations substantially change the appearance of both the product surface and the same defect category. All images were annotated by six professional annotators recruited and organized by us, using bounding boxes for seven defect categories: \emph{scratch}, \emph{leakage}, \emph{fold}, \emph{bubble}, \emph{worn}, \emph{depressed}, and \emph{flare}. More details can be found in Appendix \ref{sec:BottleSKUappendix}.

\begin{table*}
\center
\caption{Comparison of public datasets for surface defect detection. Appearance complexity is qualitatively assessed according to surface texture, background pattern, object geometry, material reflectance, and intra-class appearance variation.}
\begin{center}
\resizebox{\linewidth}{!}{ 
\begin{tabular}{llccccc}
\hline
\textbf{Dataset} & \textbf{Application} & \textbf{Number} & \textbf{ Category} & \textbf{Appearance Complexity} & \multicolumn{1}{l}{\textbf{Inspection Level}} & \textbf{Image Resolution Pixels} \\ \hline
NEU-DET~\cite{song2013neu} & Hot-rolled steel strip & 1,800 & 6 & Low & Local & $120\times120$ \\
GC10-DET~\cite{lv2020gc10det} & Steel strip & 2,294 & 10 & Medium & Local & $2048\times 1000$ \\
PVMulti-Defect~\cite{li2023gbhyolov5} & Solar photovoltaic panels & 307 & 5 & Medium & Local & $5800\times3504$ \\
HRIPCB~\cite{huang2020hripcb} & HRIPCB & 1,386 & 6 & High & Full object & $600\times600$ \\
Severstal~\cite{severstal2019steel} & Steel strip & 18,074 & 4 & Low & Local & $1600\times256$ \\
DAGM~\cite{wieler2007dagm} & Textured background & 16,100 & 6 & High & Local & $800\times800$ \\
MPCG~\cite{tao2018mpcg} & Mobile phone glass & 11,808 & - & Medium & Local & - \\
MT~\cite{huang2020magnetic} & Magnetic tile & 329 & 5 & Low & Local & - \\
APDDD~\cite{alibaba2023apddd} & Aluminum & 3,005 & 10 & Medium & Local & $2560\times1920$ \\
MVTec AD~\cite{bergmann2019mvtec} & Industrial products & 5,354 & - & Varies & Full object & $700\times700$ // $1024\times1024$ \\
Ours & Bottle surface & 13,119 & 7 & High & Full object & $1240\times1624$ mainly \\ \hline
\end{tabular}
} 
\label{datasetlist} 
\end{center}
\end{table*}

\section{Experiments} 

\subsection{Datasets and Evaluation Metrics}
\label{subsec:datasets-and-metrics}

\paragraph{Bottle-SKU.}
Bottle-SKU is a full-object shampoo-bottle surface-defect benchmark with 13,119 images, 20,419 instance-level bounding boxes, seven defect categories, and 48 real product SKUs. We use a SKU-disjoint hold-out split rather than a random image split. The training set contains 10,495 images and 16,194 defect instances from 38 SKUs, whereas the test set contains 2,624 images and 4,225 instances from the remaining 10 SKUs. Therefore, no product SKU is shared between the two partitions, while all seven defect categories occur in both. This protocol evaluates whether a detector can recognize shared defect semantics under the appearance changes introduced by unseen bottle variants. 

\paragraph{GC10-DET-SKU.}

GC10-DET~\cite{lv2020gc10det} is a public high-resolution metallic-surface defect dataset with ten defect categories, comprising 2,292 images of resolution $2048\times1000$ and 3,563 bounding boxes. GC10-DET does not provide a predefined train/test split. We therefore construct GC10-DET-SKU by clustering features from all usable images and holding out complete clusters as pseudo-SKUs. For each image, we extract a 768-dimensional L2-normalized CLS feature from a ViT-B/16 model pretrained on ImageNet-21k, and cluster the features using K-means with a fixed seed of 3407. We search $K\in\{4,\ldots,12\}$ and select $K=7$, which gives the highest cosine-silhouette score among partitions satisfying the minimum-cluster-size, class-coverage, and test-ratio constraints. Two complete clusters (pseudo-SKUs SKU-1 and SKU-2) are held out for testing, while the other five clusters (SKU-3--SKU-7) form training. This yields 1,773 training images with 2,762 boxes and 519 test images with 801 boxes (a 77.4\%/22.6\% image split).  

\paragraph{Evaluation Metrics.}
We use the standard COCO bounding-box evaluation protocol for all methods. The primary metric is AP, namely the mean AP averaged over IoU thresholds from 0.50 to 0.95 with a step of 0.05. We additionally report $\rm{AP50}$, $\rm{AP75}$, and AP for small, medium, and large objects ($\rm{AP}_{\rm{S}}$, $\rm{AP}_{\rm{M}}$, and $\rm{AP}_{\rm{L}}$). For an interpretable fixed operating point, we also report precision, recall, and F1 at a confidence threshold of 0.25 and an IoU threshold of 0.50. All results are reported as percentages, and all compared methods are evaluated  nder the same dataset configuration. Predictions are processed in descending confidence order and matched one-to-one with ground-truth boxes of the same class. Thus, with true positives (TP), false positives (FP), and false negatives (FN), 
\begin{equation}
P=\frac{\rm{TP}}{\rm{TP}+\rm{FP}},\qquad
R=\frac{\rm{TP}}{\rm{TP}+\rm{FN}},\qquad
F_1=\frac{2PR}{P+R}.
\end{equation}
The fixed-threshold $P$, $R$, and $F_1$ characterize an operating point and
are reported separately from COCO mAP.

\subsection{Comparison on Public Benchmark Datasets}

Table~\ref{tab:main-results} compares ProDMM with representative lightweight and vision-language-based detectors. The parameter column denotes the total number of model parameters. 

\paragraph{Results on Bottle-SKU.}
On the SKU-disjoint Bottle-SKU test set, ProDMM obtains the highest overall AP (16.0), AP50 (40.8), and AP75 (8.9) among the evaluated methods. The strongest baseline scores on these three measures are 14.9, 38.9, and 8.0, respectively, giving gains of 1.1, 1.9, and 0.9 percentage points. The improvement at AP75 is particularly relevant because the defects frequently occupy only a small portion of a full bottle image and require accurate localization under previously unseen product appearances. At the fixed operating point, ProDMM also achieves the highest recall (67.7), showing that the model recovers more annotated defects. Its precision (38.3) and F1 (49.0) are lower than those of several comparison methods, which indicates a precision--recall trade-off at the fixed score threshold rather than a contradiction with the ranking by COCO AP. These threshold-dependent measures should therefore be interpreted together with the full precision--recall curve summarized by AP.

\paragraph{Results on GC10-DET-SKU.}
On the generated GC10-DET-SKU held-out protocol, ProDMM reaches 37.8 AP,
72.1 AP50, and 37.2 AP75. These values exceed the strongest baseline results
by 2.0, 4.0, and 2.2 percentage points, respectively. ProDMM further obtains
72.0 recall and an F1 score of 65.8. The latter is the highest among the
vision-language baselines in Table~\ref{tab:main-results}. The result on this
public metallic-surface dataset suggests that prototype-conditioned modulation
is not restricted to the bottle imagery. Since GC10-DET-SKU is a constructed
pseudo-domain protocol, this comparison is a controlled
cross-domain evaluation on a shared split.

\begin{table*}[]
\caption{Comparison on Bottle-SKU and the GC10-DET-SKU pseudo-domain
protocol. All values are percentages.}
\label{tab:main-results}
\resizebox{\textwidth}{!}{
\begin{tabular}{llllllllllllllll}
\hline
\multicolumn{1}{l|}{\multirow{2}{*}{\textbf{Index}}} & \multirow{2}{*}{\textbf{Algorithm}} & \multirow{2}{*}{\textbf{Publish}} & \multicolumn{6}{c}{\textbf{Bottle-SKU}} & \multicolumn{6}{c}{\textbf{GC10-DET-SKU}} & \multirow{2}{*}{\textbf{Params}} \\ \cline{4-15}
\multicolumn{1}{l|}{} &  &  & AP & AP50 & AP75 & P & R & F1 & AP & AP50 & AP75 & P & R & F1 &  \\ \hline
\multicolumn{16}{c}{\textbf{Lightweight Detector}} \\ \hline
\multicolumn{1}{l|}{1} & Faster R-CNN~\cite{ren2017faster} & TPAMI 2017 & 9.6 & 28.2 & 3.6 & 32.6 & 60.4 & 42.3 & 20.8 & 43.6 & 19.0 & 29.7 & 60.4 & 39.8 & 41.38M \\
\multicolumn{1}{l|}{2} & Cascade R-CNN~\cite{cai2018cascade} & CVPR 2018 & 10.6 & 29.8 & 4.7 & 42.2 & 55.9 & 48.1 & 26.0 & 51.5 & 23.1 & 44.5 & 62.6 & 52.0 & 69.17M \\
\multicolumn{1}{l|}{3} & FCOS~\cite{tian2019fcos} & ICCV 2019 & 9.5 & 28.2 & 4.1 & 56.7 & 40.9 & 47.5 & 18.3 & 41.7 & 15.1 & 59.3 & 35.3 & 44.2 & 32.13M \\
\multicolumn{1}{l|}{4} & GFL~\cite{li2020generalized} & NeurIPS 2020 & 11.1 & 31.3 & 5.0 & 30.4 & 63.2 & 41.1 & 24.8 & 50.6 & 23.0 & 31.4 & 61.2 & 41.5 & 32.27M \\
\multicolumn{1}{l|}{5} & Deformable DETR~\cite{zhu2021deformable} & ICLR 2021 & 11.0 & 31.3 & 4.9 & 38.0 & 56.4 & 45.4 & 29.0 & 57.5 & 27.0 & 49.2 & 65.1 & 56.1 & 40.10M \\
\multicolumn{1}{l|}{6} & TOOD~\cite{feng2021tood} & ICCV 2021 & 11.2 & 31.8 & 4.4 & 38.1 & 59.4 & 46.4 & 21.5 & 44.1 & 19.4 & 44.1 & 49.8 & 46.8 & 32.03M \\
\multicolumn{1}{l|}{7} & YOLOX-S~\cite{ge2021yolox} & arXiv 2021 & 6.3 & 18.7 & 2.5 & 29.7 & 43.2 & 35.2 & 23.4 & 50.3 & 19.5 & 51.3 & 51.3 & 51.3 & 8.94M \\
\multicolumn{1}{l|}{8} & RTMDet-X~\cite{lyu2022rtmdet} & arXiv 2022 & 6.1 & 17.1 & 2.6 & 20.0 & 48.7 & 28.4 & 27.6 & 54.8 & 27.0 & 32.2 & 65.0 & 43.0 & 94.78M \\
\multicolumn{1}{l|}{9} & DINO~\cite{zhang2023dino} & ICLR 2023 & 11.0 & 31.3 & 5.0 & 39.9 & 56.2 & 46.7 & 27.6 & 52.6 & 26.4 & 49.8 & 50.4 & 50.1 & 47.55M \\
\multicolumn{1}{l|}{10} & RT-DETR-L~\cite{zhao2024rtdetr} & CVPR 2024 & 7.9 & 24.8 & 2.7 & 36.3 & 31.3 & 33.6 & 29.4 & 59.7 & 27.6 & 66.9 & 60.6 & 63.6 & 32.82M \\
\multicolumn{1}{l|}{11} & RT-DETR-X~\cite{zhao2024rtdetr} & CVPR 2024 & 7.7 & 24.6 & 2.9 & 36.1 & 35.6 & 35.8 & 27.0 & 58.9 & 20.7 & 66.9 & 59.0 & 62.7 & 67.32M \\
\multicolumn{1}{l|}{12} & YOLO26-N~\cite{ultralytics2026yolo26} & arXiv 2026 & 11.2 & 30.0 & 5.5 & 43.5 & 35.9 & 39.4 & 29.3 & 56.0 & 29.4 & 63.7 & 56.4 & 59.8 & 2.51M \\
\multicolumn{1}{l|}{13} & YOLO26-S~\cite{ultralytics2026yolo26} & arXiv 2026 & 12.0 & 31.4 & 5.7 & 44.4 & 34.4 & 38.8 & 28.9 & 55.8 & 26.3 & 67.9 & 54.0 & 60.2 & 9.95M \\
\multicolumn{1}{l|}{14} & YOLO26-M~\cite{ultralytics2026yolo26} & arXiv 2026 & 12.1 & 31.6 & 5.8 & 40.3 & 36.9 & 38.5 & 29.6 & 55.9 & 27.2 & 61.4 & 61.7 & 61.5 & 21.78M \\
\multicolumn{1}{l|}{15} & YOLO26-L~\cite{ultralytics2026yolo26} & arXiv 2026 & 12.1 & 31.8 & 5.8 & 43.8 & 34.0 & 38.3 & 29.4 & 58.1 & 23.4 & 63.1 & 58.6 & 60.8 & 26.19M \\
\multicolumn{1}{l|}{16} & YOLO26-X~\cite{ultralytics2026yolo26} & arXiv 2026 & 12.7 & 33.7 & 6.0 & 41.9 & 38.3 & 40.0 & 28.9 & 26.4 & 27.0 & 65.7 & 58.8 & 62.1 & 58.82M \\
\multicolumn{1}{l|}{17} & YOLOv8-M~\cite{jocher2023ultralytics} & 2023 & 11.9 & 32.1 & 5.8 & 37.7 & 36.2 & 37.0 & 28.0 & 57.2 & 24.8 & 62.8 & 57.8 & 60.2 & 25.84M \\ \hline
\multicolumn{16}{c}{\textbf{VLM-based}} \\ \hline
\multicolumn{1}{l|}{18} & WeDetect~\cite{fu2026wedetect} & CVPR 2026 & 13.9 & 34.7 & 7.5 & 53.4 & 52.2 & 52.8 & 32.8 & 62.8 & 30.4 & 53.7 & 67.9 & 59.9 & 386.79M \\
\multicolumn{1}{l|}{19} & DOSOD~\cite{he2024dosod} & arXiv 2024 & 13.1 & 33.7 & 6.6 & 49.9 & 53.5 & 51.6 & 28.6 & 55.2 & 26.8 & 56.4 & 61.1 & 58.7 & 108.24M \\
\multicolumn{1}{l|}{20} & PET-DINO~\cite{fu2026pet} & CVPR 2026 & 14.6 & 38.9 & 7.0 & 41.4 & 64.8 & 50.5 & 32.3 & 61.3 & 29.9 & 46.3 & 69.7 & 55.7 & 176.15M \\
\multicolumn{1}{l|}{21} & MM-GDINO-T~\cite{chen2024mmgroundingdino} & arXiv 2024 & 14.9 & 38.2 & 8.0 & 39.8 & 65.8 & 49.6 & 35.8 & 68.1 & 35.0 & 49.7 & 73.5 & 59.3 & 172.91M \\
\multicolumn{1}{l|}{22} & YOLO-World~\cite{cheng2024yoloworld} & CVPR 2024 & 13.1 & 32.9 & 6.7 & 55.7 & 49.5 & 52.5 & 31.3 & 59.8 & 30.0 & 65.6 & 56.3 & 60.6 & 110.26M \\
\multicolumn{1}{l|}{23} & Ours & - & 16.0 & 40.8 & 8.9 & 38.3 & 67.7 & 49.0 & 37.8&72.1&37.2& 60.5 & 72.0 & 65.8 & 252.68M \\ \hline
\end{tabular}
}
\end{table*}

\subsection{Ablation Study}
\paragraph{Component analysis.}
Table~\ref{tab:component-ablation} isolates the contributions of the three
main components. Starting from the frozen GroundingDINO baseline, adding
dynamic modulation without prototypes raises AP from 15.1 to 15.5. Conditioning
the modulation on SKU prototypes provides a further 0.3-point AP gain and
improves AP50 from 40.7 to 41.6, showing that the prototype representation is
useful beyond an image-only dynamic update. Adding Defect-DCA yields the best
overall AP of 16.0 and increases AP75 from 8.0 to 8.9. Although AP50 changes
from 41.6 to 40.8, the gains in overall AP and stricter-localization AP75
indicate that DCA changes the score and localization trade-off rather than
uniformly improving every operating point.

\begin{table}[]
\caption{Component Analysis.}
\label{tab:component-ablation}

\begin{center}
\begin{tabular}{c|ccc|c|c|c|c|c|c}
\hline
\textbf{No.} & \textbf{Prototype} & \textbf{DMT} & \multicolumn{1}{l|}{\textbf{DCA}} & \multicolumn{1}{l|}{\textbf{AP}} & \multicolumn{1}{l|}{\textbf{AP$_{50}$}} & \multicolumn{1}{l|}{\textbf{AP$_{75}$}} & \multicolumn{1}{l|}{\textbf{AP$_{S}$}} & \multicolumn{1}{l|}{\textbf{AP$_{M}$}} & \multicolumn{1}{l}{\textbf{AP$_{L}$}} \\ \hline
01 & \no & \no & \no & 15.1 & 39.8 & 7.9 & 2.5 & 9.1 & 16.6 \\
02 & \no & \yes & \no & 15.5 & 40.7 & 8.1 & 2.2 & 8.9 & 17.1 \\
03 & \yes & \yes & \no & 15.8 & 41.6 & 8.0 & 2.2 & 8.4 & 17.4 \\
04 & \yes & \yes & \yes & 16.0&40.8&8.9&2.3&8.7&17.9 \\ \hline
\end{tabular}
\end{center}
\end{table}

\paragraph{Modulation width.}
Table~\ref{tab:modulation-width} varies the latent width used to generate the QLer and VLer updates. The 256-dimensional setting achieves the best overall AP, AP50, and AP75. Smaller widths retain competitive results but do not provide a monotonic advantage. For example, the 64-dimensional setting has a higher AP$_S$ but lower overall AP. This behavior is consistent with a capacity trade-off. The generator must encode both shared defect semantics and SKU-dependent appearance cues, while excessive compression can limit the fidelity of the dynamic updates. 
\begin{table}[t]
\centering
\label{tab:ablation}
\begin{minipage}[t]{0.5\linewidth}
\centering
\caption{Analysis of Layer Dimension in QLer and VLer.}
\label{tab:modulation-width}
\resizebox{\linewidth}{!}{
\begin{tabular}{c|c|c|c|c|c|c|c}
\hline
\textbf{No.} & \textbf{Dimension} & \textbf{AP} & \textbf{AP$_{50}$} & \textbf{AP$_{75}$} & \textbf{AP$_{S}$} & \textbf{AP$_{M}$} & \textbf{AP$_{L}$} \\ \hline
01 & 16  & 15.5 & 39.9 & 8.1 & 2.7 & 9.9  & 16.9 \\
02 & 64  & 15.0 & 39.0 & 7.9 & 3.8 & 10.5 & 16.6 \\
03 & 128 & 15.5 & 39.8 & 8.5 & 2.6 & 10.7 & 17.1 \\
04 & 256 & 16.0 & 40.8 & 8.9 & 2.3 & 8.7  & 17.9 \\ \hline
\end{tabular}
}
\end{minipage}
\hfill
\begin{minipage}[t]{0.46\linewidth}
\centering
\caption{Analysis of the Number of Extracted Features for Prototypes.}
\label{tab:prototype-sampling}
\resizebox{\linewidth}{!}{
\begin{tabular}{c|c|c|c|c|c|c|c}
\hline
\textbf{No.} & \textbf{Features} & \textbf{AP} & \textbf{AP$_{50}$} & \textbf{AP$_{75}$} & \textbf{AP$_{S}$} & \textbf{AP$_{M}$} & \textbf{AP$_{L}$} \\ \hline
01 & 100 & 16.0 & 40.8 & 8.9 & 2.3 & 8.7  & 17.9 \\
02 & 300 & 15.8 & 40.5 & 8.4 & 2.9 & 9.6  & 17.5 \\
03 & 500 & 15.5 & 40.2 & 8.2 & 2.7 & 10.1 & 17.1 \\ \hline
\end{tabular}
}
\end{minipage}

\end{table}


\paragraph{Prototype support size.}
Table~\ref{tab:prototype-sampling} studies the number of training-image features used to form each SKU prototype. Increasing the support size from 100 to 300 results in a modest 0.2-point AP decrease, whereas using 500 features does not yield further gains. The small spread across these settings suggests that the prototype estimate becomes stable with a limited number of samples. A very large support set is therefore unnecessary for obtaining a useful domain feature.


\section{Conclusion} 

We introduced Bottle-SKU, a full-object shampoo-bottle surface-defect benchmark designed to evaluate detection under product-variant appearance shifts. The dataset contains 13,119 images, 20,419 instance-level bounding
boxes, seven defect categories, and 48 SKUs. Its SKU-disjoint protocol trains on 38 product variants and evaluates on 10 unseen variants, separating shared
defect recognition from memorization of a particular product appearance. We further presented Prototype-Guided Dynamic Domain Modulation (ProDMM). ProDMM summarizes training images with offline SKU prototypes, predicts a soft product-domain distribution for each input and uses it to generate dynamic residual updates for the Query and Value projections of the frozen backbone. Experiments on the Bottle-SKU and GC10-DET-SKU datasets demonstrate that prototype-conditioned modulation is suitable for the multi-SKU task. Component studies show complementary benefits from dynamic modulation, prototype conditioning, and DCA, while the prototype-support and modulation-width experiments indicate that the method does not require a large per-SKU image support set to be effective.

\subsection*{AI use statement}
In this work, we used generative AI tools to improve the quality of written English and for stylistic refinement, with attentive supervision by the authors. No scientific content, experimental design or data analysis were generated through LLMs. All AI-assisted writing was reviewed for correctness by 2 authors.



\bibliography{reference}
\bibliographystyle{iclr2027_conference}

\appendix

\section{Extended Related Work}
\label{sec:more-related-work}

\subsection{Vision-Language and Open-Vocabulary Object Detection}
Vision-language pre-training aligns visual representations with natural-language concepts. CLIP~\cite{radford2021clip} demonstrates that contrastive learning over web-scale image-text pairs can produce transferable embeddings whose textual prompts act as a flexible classifier. GLIP~\cite{li2022glip} extends this paradigm from image-level matching to region-level grounding, showing that phrases can supervise both semantic recognition and localization. Caption-driven open-vocabulary detection uses image-level descriptions to transfer supervision beyond the detector label set~\cite{zareian2021ovd}, while Detic~\cite{gu2022detic} demonstrates that image-level supervision can scale detection to tens of thousands of categories through a large vocabulary classifier.

Transformer detectors form an effective backbone for language-conditioned localization. Deformable DETR~\cite{zhu2021deformable} replaces dense global attention with reference-point sampling over multi-scale features, making transformer detection practical for small objects and high-resolution inputs. DINO~\cite{zhang2023dino} improves DETR convergence by combining denoising queries, improved anchor initialization, and contrastive query selection, and consequently provides a strong closed-set detection foundation. Grounding DINO~\cite{liu2024groundingdino} couples this detector design with grounded vision-language pre-training. Text tokens and image features interact in the feature enhancer and language-guided query selection, enabling open-set localization from natural-language prompts. Grounding DINO 1.5~\cite{ren2024groundingdino15} further focuses on stronger open-set accuracy and deployment-oriented efficiency.

Recent detectors explore different ways to inject language at inference. YOLO-World~\cite{cheng2024yoloworld} distills vision-language knowledge into a YOLO-style architecture and uses prompt- or vocabulary-embeddings for real-time open-vocabulary prediction. DetCLIPv3~\cite{yao2024detclipv3} treats open-vocabulary detection as a more general generative recognition problem, improving the ability to transfer from diverse textual descriptions. T-Rex2~\cite{jiang2024trex2} studies the synergy between visual exemplars and text prompts, which is useful when category names alone are insufficient. OV-DINO~\cite{wang2024ovdino} introduces language-aware selective fusion so that visual features receive category-relevant semantic information without uniformly mixing all language tokens. LLMDet~\cite{fu2025llmdet} uses large-language-model supervision to enrich detector semantics, while MM-GroundingDINO~\cite{chen2024mmgroundingdino} presents a broader grounding-and-detection pipeline. These methods mainly target category expansion. ProDMM instead addresses SKU-dependent appearance variation while retaining the grounded detection pathway.

\subsection{Prototype-Based Domain Adaptation and Conditional Modulation}
Prototype-based learning represents a class or domain with a representative embedding. Prototypical Networks~\cite{snell2017prototypical} introduced episodic learning in which a query is classified by its distance to the mean embedding of each support class, providing a simple mechanism for few-shot generalization. N-pair loss~\cite{sohn2016nce} replaces pairwise sampling with multiple negative examples in one objective, improving metric separation and reducing the instability of contrastive training. Supervised contrastive learning~\cite{khosla2020supervised} generalizes this idea by pulling together all samples sharing a label while pushing apart samples from other labels, which is useful for modeling intra-class variation. For domain shift, DANN~\cite{ganin2016dann} introduces a gradient-reversal domain classifier to learn features that confuse source and target domains. ADDA~\cite{tzeng2017adversarial} decouples source representation learning from target mapping, allowing an unlabeled target encoder to be aligned adversarially. Domain Adaptive Faster R-CNN~\cite{chen2018dafa} transfers these principles to object detection by aligning image-level and instance-level features. DomainBed~\cite{zhou2021domainbed} standardizes domain-generalization evaluation, while SWAD~\cite{cha2022domain} seeks flat minima that are empirically less sensitive to unseen domain changes.

Conditional modulation changes a representation as a function of an auxiliary signal. FiLM~\cite{perez2018film} predicts feature-wise scale and bias parameters from the conditioning input, allowing the same visual backbone to implement different computations for different contexts. CondConv~\cite{yang2019condconv} extends this idea to convolution by generating input-dependent mixtures of kernel experts. Sparsely gated MoE~\cite{shazeer2017moe} demonstrates that a learned gate can activate only a small subset of experts, increasing model capacity without proportional computation, and Switch Transformer~\cite{switch2022} simplifies this expert-selection mechanism to one expert per token for efficient large-scale training. Parameter-efficient transfer uses bottleneck adapters~\cite{houlsby2019adapter} to isolate task-specific updates, LoRA~\cite{hu2022lora} to inject trainable low-rank matrices into frozen linear layers, and AdaptFormer~\cite{chen2022adaptformer} to adapt vision transformers with lightweight residual branches.
\begin{figure}
\centering
\includegraphics[width=1\linewidth]{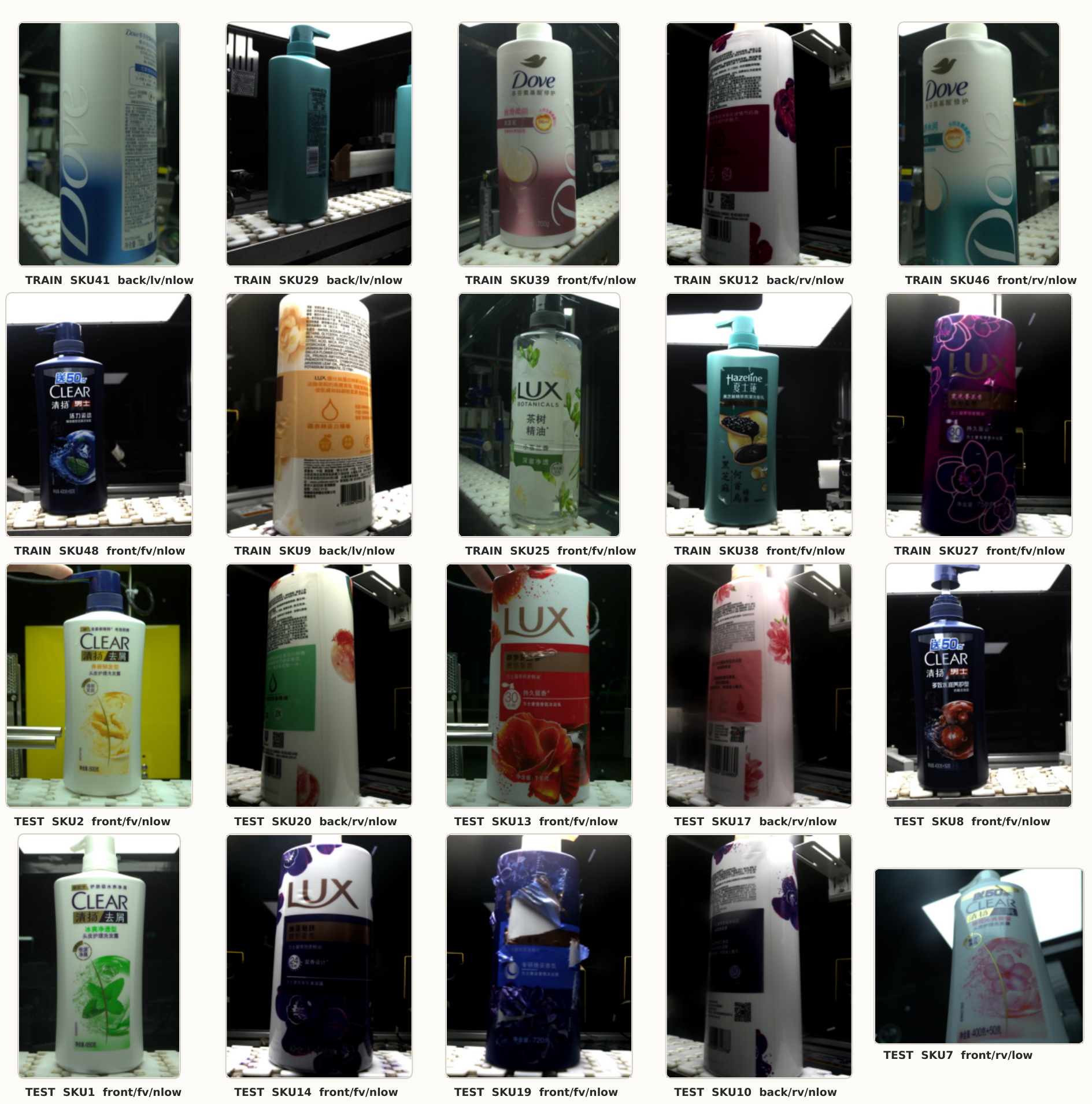}
\caption{Examples of bottle images from different SKUs in our dataset.} 
\label{fig:bottle_sku_image_samples}
\end{figure}
\section{Implementation Details} 
\label{ImpleDetails}
All experiments are conducted on NVIDIA A800 80GB GPUs and initialized from the official GroundingDINO Swin-B checkpoint. For the main frozen ProDMM setting, the Swin-B backbone and BERT encoder are frozen, while the dynamic modulation modules and detection adaptation pathway remain trainable. We build 38 offline SKU prototypes from at most 100 training images per SKU. Each prototype feature is obtained by global average pooling the final Swin stage and applying feature normalization before prototype aggregation.

The Domain Encoder uses the initial patch-embedding tokens. A $1\times1$ spatial gate is fused with the original token feature, followed by global pooling and an MLP domain classifier. The DMM encoder is a two-layer GELU MLP that projects the 1024-dimensional prototype mixture to a 256-dimensional condition. It drives 48 independent Swin Q/V heads and two 256-dimensional Defect-DCA Q/V residual heads. QLer and VLer use rank $r=1$ and are implemented with low-rank down/up projections, biases, and a scalar modulation factor. The dynamic output heads use zero-output initialization with a fixed adapter basis, so the initial residual update is zero.

Defect-only DCA uses eight attention heads over four 256-dimensional neck levels. The SKU-conditioned spatial prior is represented by a $16\times16$ grid for each domain, attention head, and neck level. The soft SKU posterior mixes these grids, which are bilinearly interpolated to the corresponding attention resolution. $\mathcal{L}_{reg}$ is the mean-square penalty over the generated QLer/VLer increments, DCA Q/V residuals, and spatial-prior parameters. We use $\beta=1.0$ and $\gamma=10^{-4}$. Training uses AdamW with learning rate $10^{-4}$, weight decay $10^{-4}$, gradient clipping at 0.1, a constant learning rate, and a fixed seed of 3407. We train for 30 epochs with training and validation batch sizes of 18 and 36, respectively.


\section{Bottle-SKU Dataset Details}
\label{sec:BottleSKUappendix}
\subsection{SKU Definition and Product Grouping}

Bottle-SKU was curated at the product-variant level rather than at the coarse brand or bottle-color level as shown in Fig. \ref{fig:bottle_sku_image_samples}. During annotation, bottles with different label descriptions or package specifications were treated as distinct product variants. Each filename contains three groups of metadata separated by underscores. These groups include a \emph{product descriptor} (\texttt{SKU-ID} and package weight), a \emph{capture descriptor} (visible side, horizontal view, and camera elevation), and a unique image identifier.

For example, \texttt{SKU3\_450g\_back\_fv\_nlow\_2925} denotes an image from the internal product family \texttt{SKU3} with a \texttt{450g} package specification. Its capture descriptor indicates the bottle back, frontal view (\texttt{fv}), and non-upward camera elevation (\texttt{nlow}). The final field \texttt{2925} is the image identifier. The first two fields jointly represent product identity, where \texttt{SKU-ID} denotes the curated product/label family and the weight distinguishes package variants within that family.

The acquisition fields are retained as metadata rather than defect labels. Specifically, \texttt{front} and \texttt{back} identify the visible bottle side. The tokens \texttt{fv}, \texttt{lv}, and \texttt{rv} represent frontal, left-oblique, and right-oblique views, respectively. The tokens \texttt{low} and \texttt{nlow} denote upward-looking and non-upward camera elevations. An additional suffix is used only when necessary to disambiguate duplicate image identifiers.

This organization preserves appearance factors that are relevant to industrial inspection. Images of the same product can differ substantially in reflections, label textures, bottle contours, and visible regions under different sides and viewpoints. Bottle-SKU therefore supports defect detection under SKU-related appearance variation while retaining the metadata needed for controlled analysis.

\subsection{Dataset Annotation}
\label{sec:BottleAnnotation}

\begin{figure}
\centering
\includegraphics[width=1\linewidth]{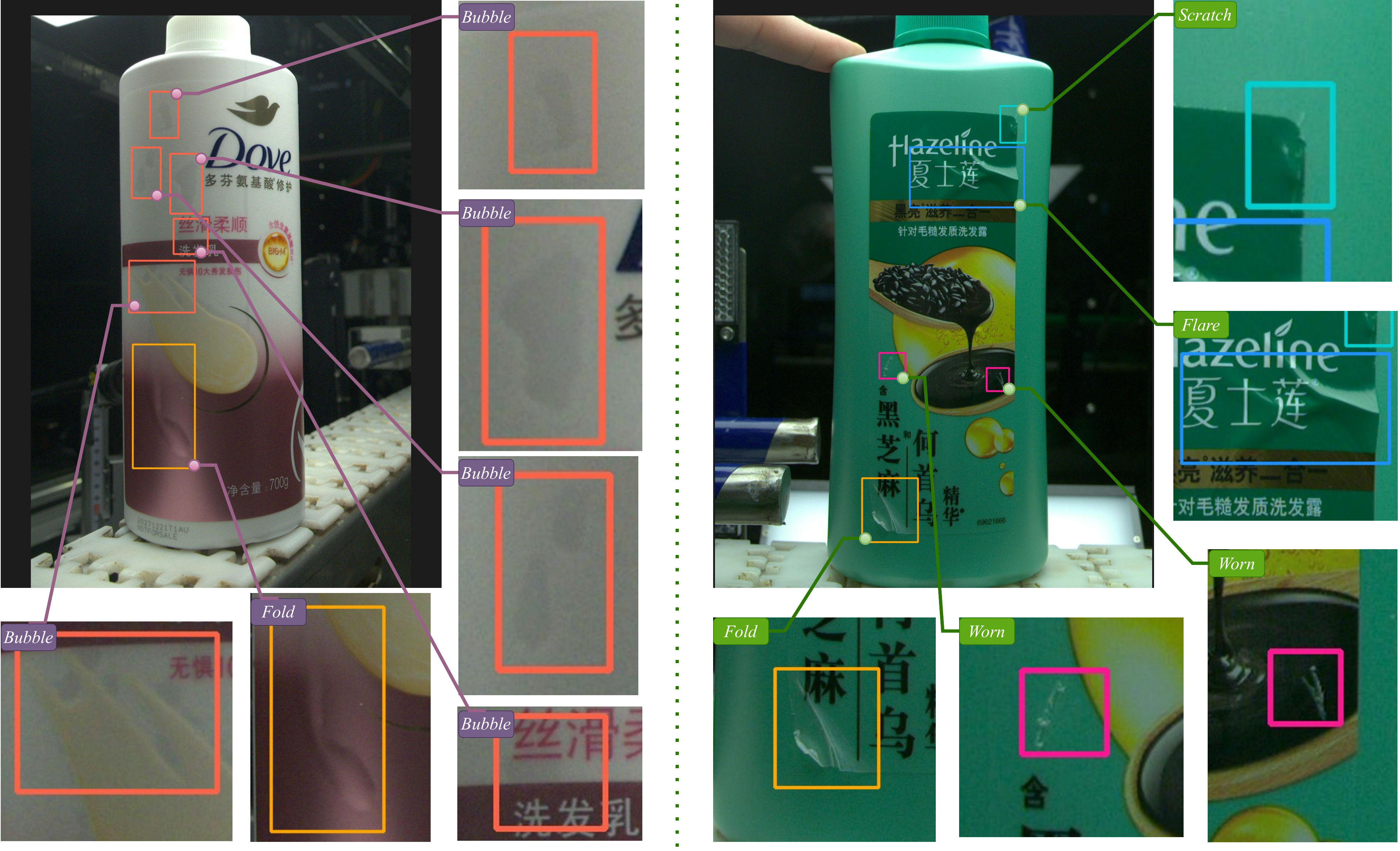}
\caption{Examples of defects appearing on bottle surfaces.} 
\label{fig:bottle_sku_defecet_present}
\end{figure}

Annotation follows the principle that a box should cover the complete visually observable defect region, rather than only the most salient local texture. The annotation samples are shown in Fig. \ref{fig:bottle_sku_defecet_present}. The detailed criteria are summarized below. 

\begin{itemize}
    \item \textbf{Scratch.} Scratches are generally elongated and strip-like. The box should cover the visible extent of the scratch.
    
    \item \textbf{Leakage.} Leakage usually exhibits a flowing pattern with irregular boundaries. A connected leakage region is annotated with one enclosing box. Separated regions are annotated independently when their connection is sufficiently weak or discontinuous.
    
    \item \textbf{Fold.} Folds often occupy a relatively broad area with uneven surfaces and blurred boundaries. The annotation covers the deformation and its associated shadow region. Folded bottle labels are also annotated as folds.
    
    \item \textbf{Bubble.} An isolated bubble is usually approximately circular and is annotated individually. Dense bubble clusters are annotated using one box covering the complete cluster rather than fragmented boxes for individual bubbles.
    
    \item \textbf{Worn.} Worn defects include missing corners, chipping, cracking, and other visibly broken regions.
    
    \item \textbf{Depressed.} Depressions are primarily revealed by local illumination and shading changes. The box covers the full depressed surface. Pixel-level boundary precision is not required.
    
    \item \textbf{Flare.} Flare refers to a misaligned or bent label that forms a cone-like hollow region, typically caused by label attachment failure.
\end{itemize}

After annotation, the labels were inspected according to the above category definitions, spatial coverage rules, and box consistency rules. In particular, annotators avoided splitting one connected defect into multiple boxes and avoided assigning a defect label solely according to bottle color or product appearance.

\subsection{Dataset Statistics}
\label{subsec:dataset-statistics}

\begin{table}[]
    \centering
    \caption{Statistics of the Bottle-SKU dataset.}
    \label{tab:bottlesku-statistics}
    \small
    \setlength{\tabcolsep}{4pt}
    \begin{tabular}{lrrrr}
        \toprule
        Split & \#SKU & \#Images & \#Instances & Ins./Img. \\
        \midrule
        Train & 38 & 10,495 & 16,194 & 1.54 \\
        Test  & 10 & 2,624  & 4,225  & 1.61 \\
        \midrule
        Total & 48 & 13,119 & 20,419 & 1.56 \\
        \bottomrule
    \end{tabular}
\end{table}

Table~\ref{tab:bottlesku-statistics} summarizes the scale and split of Bottle-SKU. The dataset contains 13,119 images, 20,419 defect instances, 48 SKU groups, and seven defect categories. We adopt a SKU-disjoint evaluation protocol. The training set contains 38 SKU groups, whereas the test set contains 10 different SKU groups.

All seven defect categories occur in both splits. The instance distribution is naturally long-tailed. \emph{Fold} is the most frequent category, accounting for 38.7\% and 47.2\% of the training and test instances, respectively, whereas \emph{flare} is comparatively scarce. This distribution reflects the practical prevalence of different bottle-surface defects and makes the task more challenging than a class-balanced benchmark.

\begin{figure}[]
    \centering
    \includegraphics[width=1\linewidth]{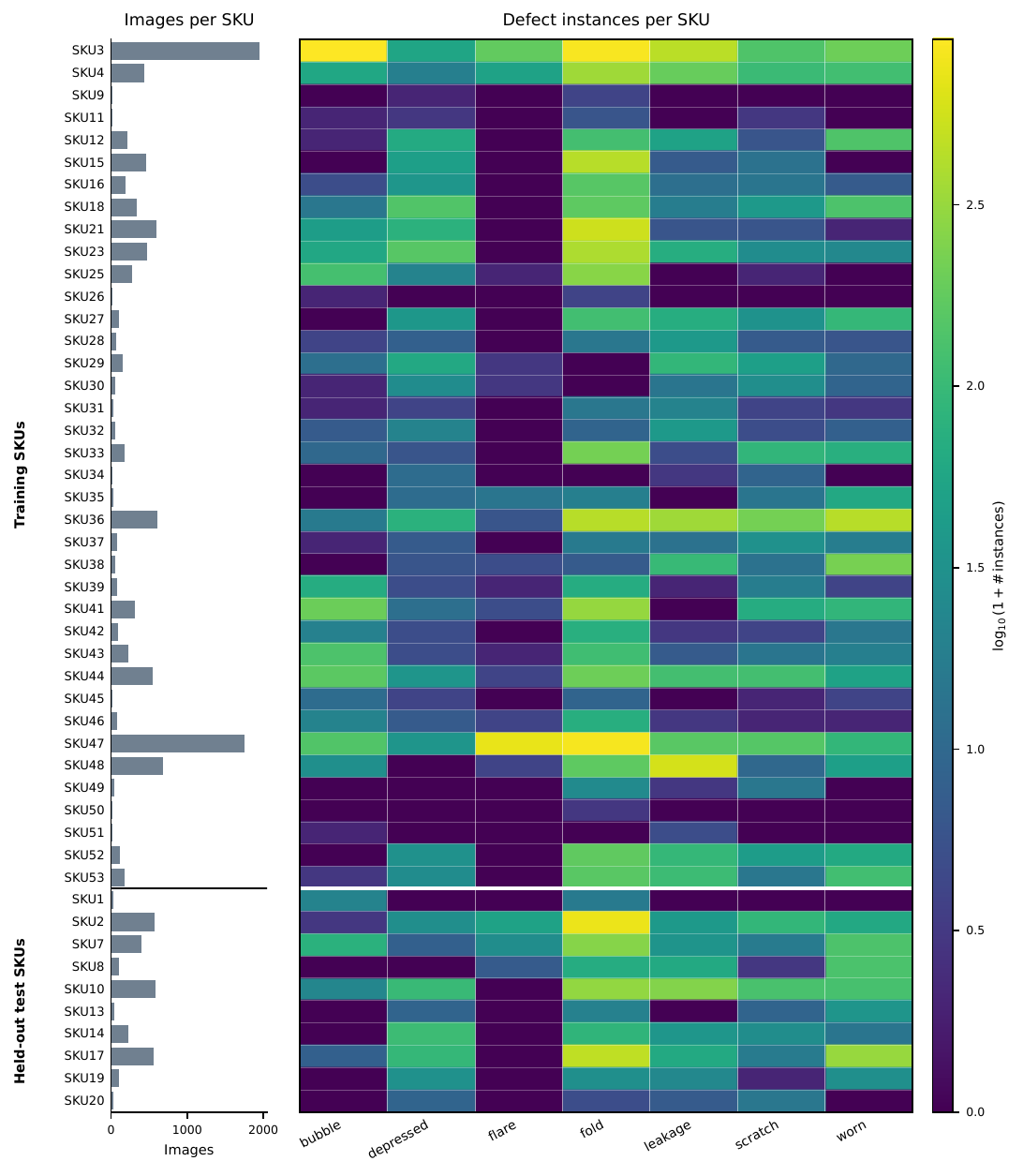}
    \caption{Distribution of annotated defect instances across Bottle-SKU
    product variants. The rows are divided into training SKUs and
    held-out test SKUs. Color intensity represents
    $\log_{10}(1+\#\mathrm{instances})$, and the bars on the left indicate
    the number of images for each SKU.}
    \label{fig:sku-defect-heatmap}
\end{figure}

Figure~\ref{fig:sku-defect-heatmap} presents the distribution of defect instances across product variants. Each row corresponds to one SKU and each column corresponds to one defect category. The color indicates the log-scaled number of annotated instances, while the bars on the left show the number of images for each SKU. The distribution is heterogeneous, with different SKUs exhibiting different defect compositions and substantially different numbers of annotated instances. In addition, the training and held-out test SKUs show different category frequencies. This confirms that the evaluation protocol involves both product-appearance variation and domain-dependent defect imbalance. 

\begin{figure}
    \centering
    \includegraphics[width=\linewidth]{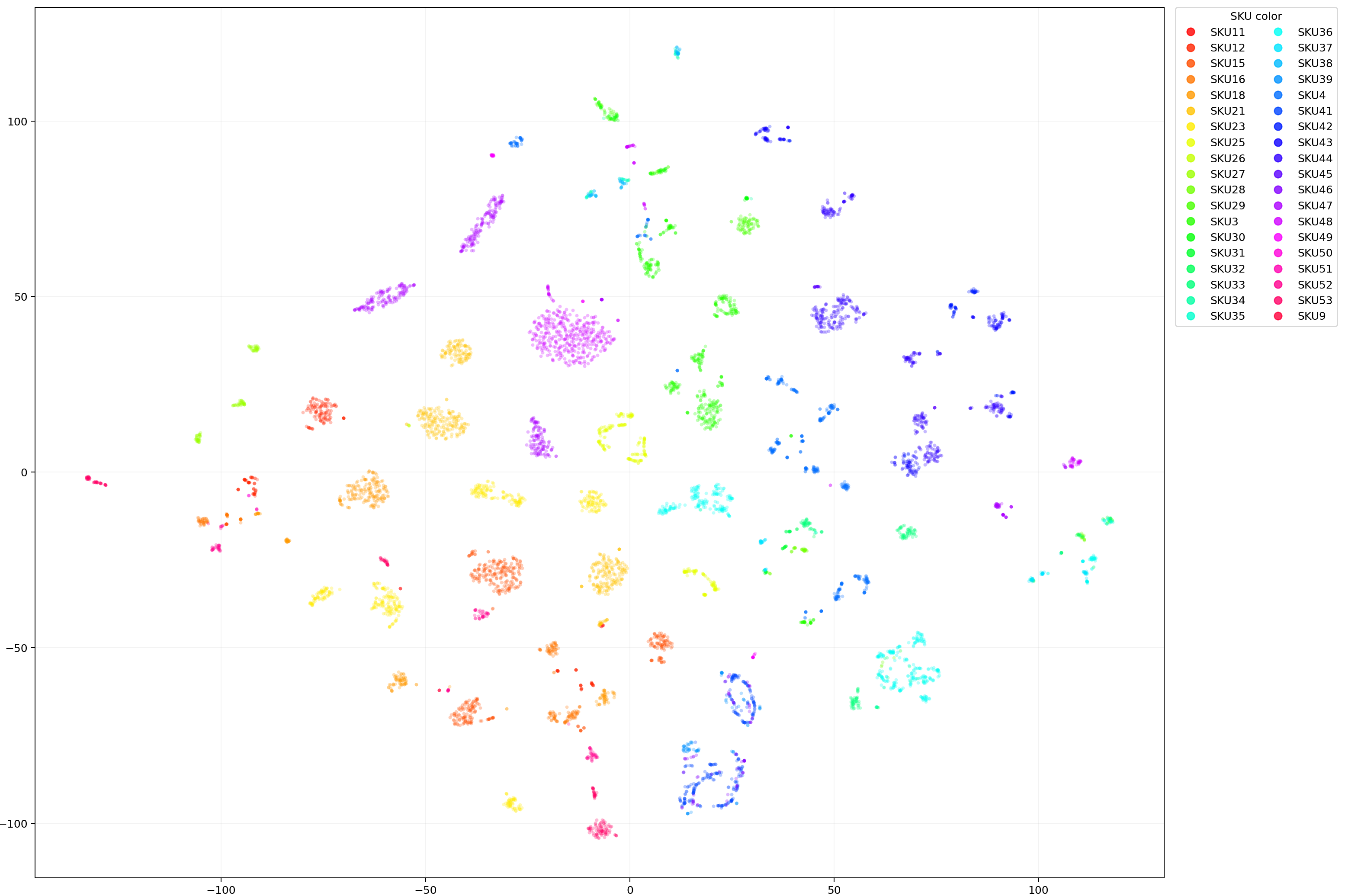}
    \caption{A t-SNE visualization of the features extracted from the training set for the 38 SKU classes.}
    \label{fig:sku-tsne}
\end{figure}

Figure~\ref{fig:sku-tsne} visualizes the frozen visual features of training images from each of the 38 training SKU groups. The figure illustrates both intra-SKU appearance variation and partial overlap between visually similar SKU groups. Such overlap is expected in real production settings, where bottles with different labels, package specifications, or viewpoints share similar colors, reflections, and local surface textures. This motivates the use of soft prototype-conditioned modulation instead of a hard SKU assignment.

\section{Additional results}

\paragraph{Comparison with static parameter-efficient updates.}
Table~\ref{tab:peft-comparison} compares the proposed prototype-conditioned
updates with static LoRA and AdaptFormer baselines under the frozen-backbone
setting. ProDMM achieves the highest AP (16.0) and AP75 (8.9), exceeding the
best LoRA result by 0.2 and 0.1 points, respectively, and AdaptFormer by 0.3
and 0.6 points. AdaptFormer obtains a higher AP$_S$, so the comparison does
not imply that ProDMM dominates every object-size regime. Instead, it supports
the central claim that conditioning the update on the inferred product domain
improves the aggregate cross-SKU detection result over a single static
adapter.

\begin{table}[]
\caption{Comparison with static parameter-efficient fine-tuning baselines.}
\label{tab:peft-comparison}
\begin{center}
\begin{tabular}{c|c|c|c|c|c|c|c}
\hline
\textbf{No.} & \textbf{Methods} & \multicolumn{1}{l|}{\textbf{AP}} & \multicolumn{1}{l|}{\textbf{AP$_{50}$}} & \multicolumn{1}{l|}{\textbf{AP$_{75}$}} & \multicolumn{1}{l|}{\textbf{AP$_{S}$}} & \multicolumn{1}{l|}{\textbf{AP$_{M}$}} & \multicolumn{1}{l}{\textbf{AP$_{L}$}} \\ \hline
01 & LoRA rank=1 & 15.5 & 39.8 & 8.1 & 2.8 & 9.8 & 17.2  \\
02 & LoRA rank=64 & 15.7 & 40.4 & 8.1 & 3.3 & 9.5 & 17.3 \\
03 & LoRA rank=128 &15.8 & 39.7 & 8.8 & 3.2 & 9.2 & 17.3  \\
04 & AdaptFormer & 15.7&40.4&8.3&3.8&9.1&17.5 \\ 
05 & Ours & 16.0&40.8&8.9&2.3&8.7&17.9\\ \hline
\end{tabular}
\end{center}
\end{table}

\paragraph{Domain loss weight.}
Table~\ref{tab:domain-loss-weight} varies the weight of the SKU loss. Increasing this weight improves AP from 15.4 at $\beta=0.25$ to 16.0 at $\beta=1.0$, and improves AP75 from 7.6 to 8.9. AP50 is not monotonic and is largest at $\beta=0.25$, again indicating that the auxiliary objective controls a localization and score-calibration trade-off. The result confirms that domain supervision must be sufficiently strong for the Domain Encoder to provide useful conditioning, while detection loss remains necessary to preserve defect-level discrimination. 

\begin{table}[]
    \caption{Parameter Analysis of the Domain Classifier Loss Weight.~}
    \label{tab:domain-loss-weight}

\begin{center}
    \begin{tabular}{c|c|c|c|c|c|c|c}
    \hline
    \textbf{No.} & \textbf{Methods} & \multicolumn{1}{l|}{\textbf{AP}} & \multicolumn{1}{l|}{\textbf{AP$_{50}$}} & \multicolumn{1}{l|}{\textbf{AP$_{75}$}} & \multicolumn{1}{l|}{\textbf{AP$_{S}$}} & \multicolumn{1}{l|}{\textbf{AP$_{M}$}} & \multicolumn{1}{l}{\textbf{AP$_{L}$}} \\ \hline
    01 & 0.25 & 15.4&41.0&7.6&2.0&8.8&17.1 \\ 
    02 & 0.5 & 15.6 & 39.9   & 8.4  & 3.0 & 10.0 & 17.2\\ 
    03 & 1.0 & 16.0&40.8&8.9&2.3&8.7&17.9 \\
    \hline
    \end{tabular}
\end{center}
    \end{table}

\section{Visualization} 

\begin{figure}
    \centering
    \includegraphics[width=1\linewidth]{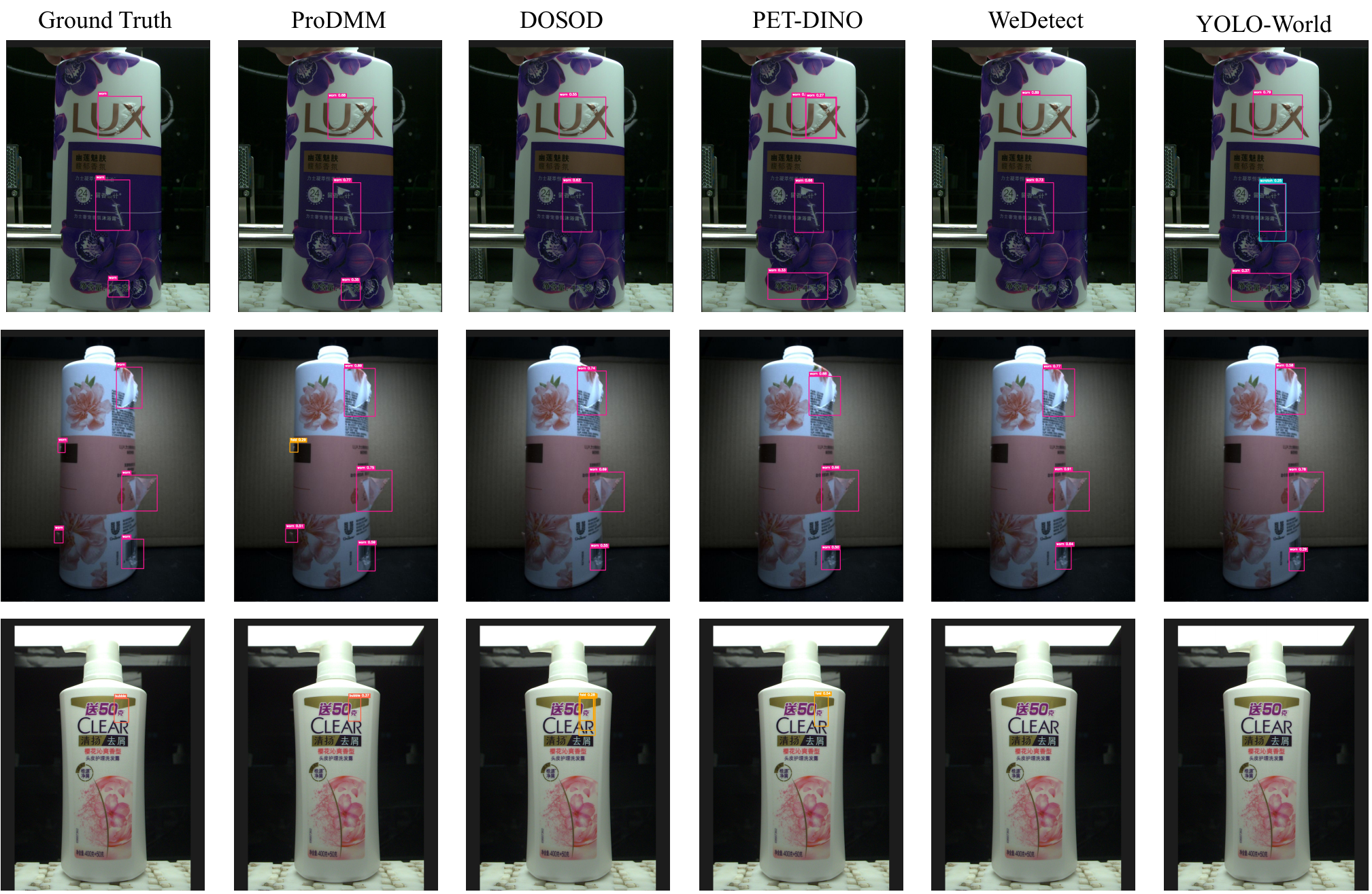}
    \caption{Comparison of Ground Truth, ProDMM, DOSOD, PET-DINO, WeDetect and YOLO-World.}
    \label{fig:method_compare}
\end{figure}

Figure~\ref{fig:method_compare} provides qualitative comparisons on Bottle-SKU testing images. All results use the same original-image geometry, class-color convention, and box-drawing rule, so differences reflect model predictions. In both examples, ProDMM detect multiple spatially separated annotated defects, including regions with weak contrast against labels or bottle surfaces. The comparison methods also detect several prominent instances, but miss or localize some of the smaller and visually ambiguous regions in these examples. 

The visualization is intended to complement rather than replace the aggregate metrics. ProDMM is not error-free. Extra or imperfectly localized predictions can occur, consistent with its fixed-threshold precision-recall trade-off in Table~\ref{tab:main-results}. Together with the quantitative results, these cases illustrate the practical role of SKU-conditioned modulation. It preserves detection of subtle defect evidence while adapting to changes in bottle appearance and imaging conditions. 

\section{Limitation Analysis} 

Despite the consistent improvements over the evaluated baselines, several limitations remain. First, ProDMM relies on an offline prototype table and on the availability of reliable SKU assignments in the training data. A single mean prototype is a compact summary of a product variant, but it cannot fully represent multimodal appearance factors such as front and back labels, viewpoint changes, illumination, or different production batches. The soft SKU representation can combine several prototypes to reduce hard assignment errors, yet its representation is still bounded by the coverage and quality of the stored prototypes. A newly introduced product, a substantially redesigned label, or strong prototype drift may therefore require prototype updating or an additional unknown-domain mechanism. 

Second, the main setting freezes the Swin-B visual backbone and the BERT text encoder. This design isolates the effect of conditional modulation and limits the number of optimized pretrained parameters, but it also restricts the adaptation capacity under severe changes in material, camera configuration, or defect morphology. ProDMM modifies only the Query and Value pathways in the visual transformer and uses residual conditioning for defect-text interaction. It does not adapt every representation in the detector. The total model still contains the large pretrained GroundingDINO parameters. Dynamic parameter generation and DCA introduce additional memory and inference computation. Thus, the frozen-backbone setting should be viewed as conditional adaptation with a reduced optimization burden, rather than as a strict reduction of the deployed model size or latency.

Third, the current evaluation is limited to one factory dataset and one constructed public pseudo-domain protocol. Bottle-SKU provides a realistic SKU-disjoint split, but it represents one product family and one acquisition pipeline. GC10-DET-SKU is derived from GC10-DET by unsupervised visual clustering and is useful for controlled cross-domain analysis. Future evaluation on independently collected factories, unseen production batches, and temporally separated test sets is needed to quantify deployment robustness. Finally, the reported precision, recall, and F1 values correspond to one confidence threshold, while the class distribution is imbalanced. Threshold calibration, cost-sensitive evaluation, and defect-specific operating points are therefore important before making production-level claims.

\end{document}

%% file: math_commands.tex
\usepackage{amsmath,amsfonts,bm}

\def\eqref#1{equation~\ref{#1}}

\def\1{\bm{1}}

\DeclareMathAlphabet{\mathsfit}{\encodingdefault}{\sfdefault}{m}{sl}
\SetMathAlphabet{\mathsfit}{bold}{\encodingdefault}{\sfdefault}{bx}{n}

